\def\year{2020}\relax
\documentclass[letterpaper]{article} 
\usepackage{aaai20}  
\usepackage{times}  
\usepackage{helvet} 
\usepackage{courier}  
\usepackage[hyphens]{url}  
\usepackage{graphicx} 
\usepackage{graphicx}  
\usepackage{amsmath}
\usepackage{enumitem}
\usepackage{makecell}
\usepackage{subcaption}
\usepackage{tabularx}
\usepackage{multirow}
\usepackage{textcomp}

\newcommand{\citet}[1]
{\citeauthor{#1}~\shortcite{#1}}
\newcommand{\citep}{\cite}

\title{Attending to Entities for Better Text Understanding}
\author{Pengxiang Cheng \\ Department of Computer Science \\ The University of Texas at Austin \\ pxcheng@utexas.edu
	\And Katrin Erk \\ Department of Linguistics \\ The University of Texas at Austin \\ katrin.erk@utexas.edu
}
\begin{document}
	
\maketitle

\begin{abstract}
Recent progress in NLP witnessed the development of large-scale pre-trained language models (GPT, BERT, XLNet, etc.) based on Transformer \cite{Vaswani2017NIPS}, and in a range of end tasks, such models have achieved state-of-the-art results, approaching human performance. This clearly demonstrates the power of the stacked self-attention architecture when paired with a sufficient number of layers and a large amount of pre-training data. However, on tasks that require complex and long-distance reasoning where surface-level cues are not enough, there is still a large gap between the pre-trained models and human performance. \citet{Strubell2018EMNLP} recently showed that it is possible to inject knowledge of syntactic structure into a model through supervised self-attention. We conjecture that a similar injection of semantic knowledge, in particular, coreference information, into an existing model would improve performance on such complex problems. On the LAMBADA \citep{Paperno2016ACL} task, we show that a model trained from scratch with coreference as auxiliary supervision for self-attention outperforms the largest GPT-2 model, setting the new state-of-the-art, while only containing a tiny fraction of parameters compared to GPT-2. We also conduct a thorough analysis of different variants of model architectures and supervision configurations, suggesting future directions on applying similar techniques to other problems.
\end{abstract}

\section{Introduction}
\label{sec::intro}
When does it help performance on NLP tasks to explicitly take linguistic structure into account?  Large-scale pre-trained language models such as ELMo \cite{Peters2018NAACL}, GPT \cite{Radford2018GPT}, BERT \cite{Devlin2019NAACL}, and XLNet \cite{Yang2019xlnet} have recently achieved state-of-the-art results on a wide range of tasks. These models, mostly built on a stacked self-attention architecture as in the Transformer \cite{Vaswani2017NIPS}, are not explicitly trained to take linguistic structure into account, but they have been shown to encode some linguistic knowledge anyway, in particular knowledge related to syntactic structure \cite{Tenney:2019,Jawahar:2019}. But on some tasks that require complex and long-distance reasoning, there is still a large gap between the pre-trained models and human performance, including tasks that require knowledge of broader discourse~\citep{Paperno2016ACL} or coreference~\citep{quoref:19}, or that require identifying valid reasoning~\citep{argmining:19}. For such tasks in particular, it is interesting to test whether explicit information about linguistic structure can be helpful, how such structure should be injected, and whether it can be helpful in addition to recent language models.
\citet{Strubell2018EMNLP} investigated how to inject syntactic knowledge into a Transformer-style model by applying supervision on self-attention weights. Intriguingly, their model, which they apply to semantic role labeling, benefits from the use of ELMo embeddings, and improves over a benchmark that uses those embeddings but no supervised self-attention. 

In this paper, we consider a task that was explicitly designed to require long-distance knowledge, the LAMBADA task~\citep{Paperno2016ACL}. LAMBADA is a language modeling task on narrative text passages, where test set data points are chosen to be easily solvable for humans given a larger preceding context of several sentences, but impossible to solve for humans given only a single sentence (see Figure \ref{fig::lambada-example} for an example). In the paper that originally introduced LAMBADA, \citet{Paperno2016ACL} report model accuracies of only 7.3\%. Since then, more recent models (GPT-2) have improved performance to 63.24\% \cite{Radford2019GPT2}, while human performance is above 80\%. So the LAMBADA dataset clearly has the characteristic described above, with a large gap between pre-trained models and human performance. 

We test whether an injection of linguistic knowledge in a similar manner to \citet{Strubell2018EMNLP}, adding supervised self-attention to an existing, non-Transformer model, will improve performance on this complex task. As LAMBADA focuses on narrative texts, we hypothesize that semantic knowledge about the entities mentioned in the passage will be particularly useful for solving the task (i.e., in Figure \ref{fig::lambada-example}, it is important to know that ``you'' refers to ``Jon'' and ``he'' refers to ``Tony'' in the first two sentences to make the right prediction). We phrase entity knowledge as knowledge about the coreference chains in the passage. We find that a \textsc{BiDAF}-based \citep{Seo2017ICLR} model trained with coreference as auxiliary supervision achieves state-of-the-art performance with only a tiny fraction of the parameters of the previous best model, the largest GPT-2 model.

We further analyze the results in more detail, finding evidence that the auxiliary supervision enables the model to better capture coreference information in the text. To provide some insights on how to apply similar techniques to other problems, we experiment with different model variants to test where best to insert the supervision into the system, and we also test different types of linguistic knowledge as supervision signals.

\section{Background and Related Work}
\label{sec::related}
\paragraph{Pre-trained Language Models} There has been rapid progress on large-scale pre-trained language models that provide contextualized embeddings for downstream tasks, setting up new state-of-the-art over traditional fixed vector word embeddings like word2vec\cite{Mikolov2013NIPS} or Glove \cite{Pennington2014EMNLP}. \citet{Peters2018NAACL} first introduced ELMo, a bidirectional-LSTM language model pre-trained on the 1B Word Benchmark, achieving the best results at the time on a broad range of tasks, including
reading comprehension, semantic role labeling, coreference resolution, and many others.

With the gaining popularity of the Transformer architecture \cite{Vaswani2017NIPS} in NLP, later efforts shifted focus to pre-trained Transformer models. \citet{Radford2018GPT} introduced GPT, by pre-training a 12-layer Transformer model as a generative language model on the BooksCorpus \cite{Zhu2015ICCV}. GPT outperformed previous state-of-the-art on 9 out of the 12 tasks studied. BERT \cite{Devlin2019NAACL} enhanced the GPT model by allowing bidirectional self-attention with a new ``masked language model'' pre-training objective, and achieved even better results than GPT. GPT-2 \cite{Radford2019GPT2} directly extended GPT with 10x larger pre-training corpus and much more complex model architectures (up to 1.5 billion parameters), and set new state-of-the-art on a number of language modeling tasks. More recent models, including XLNet \cite{Yang2019xlnet} and RoBERTa \cite{Liu2019roberta}, further improved over BERT by introducing new pre-training and optimization strategies.

\paragraph{Linguistic Structure in Pre-trained Models}
The aforementioned pre-trained models do not explicitly take any linguistic structure into account, as the pre-training objective is to predict the next word, a randomly masked word, or the next sentence. While these pre-trained models achieved state-of-the-art results on many tasks, it is still largely unknown to what extent implicit knowledge of linguistic structures, such as syntactic structure or coreference, contributes to the improvement. \citet{Tenney:2019} designed a list of probing tasks to test how well the contextualized representations learned from ELMo / GPT / BERT do on some core NLP pipeline tasks, and found out that contextualized embeddings improve largely on syntactic tasks (like part-of-speech tagging and parsing) but not so much on semantic tasks (like coreference).

\citet{Strubell2018EMNLP} recently achieved state-of-the-art performance on semantic role labeling by injecting syntactic knowledge into a Transformer-style model. In their LISA model, one self-attention head is guided to learn dependency parsing via an auxiliary supervision signal that encourages each token to only attend to its syntactic parent. They also showed that such syntactically-informed self-attention can be combined with ELMo embeddings to further improve performance over a baseline with only ELMo and self-attention but no auxiliary supervision.
In this paper, we want to investigate whether linguistic knowledge of semantic structures can be injected in a similar manner.

\begin{figure}[!t]
	\centering
	\includegraphics[width=0.9\linewidth]{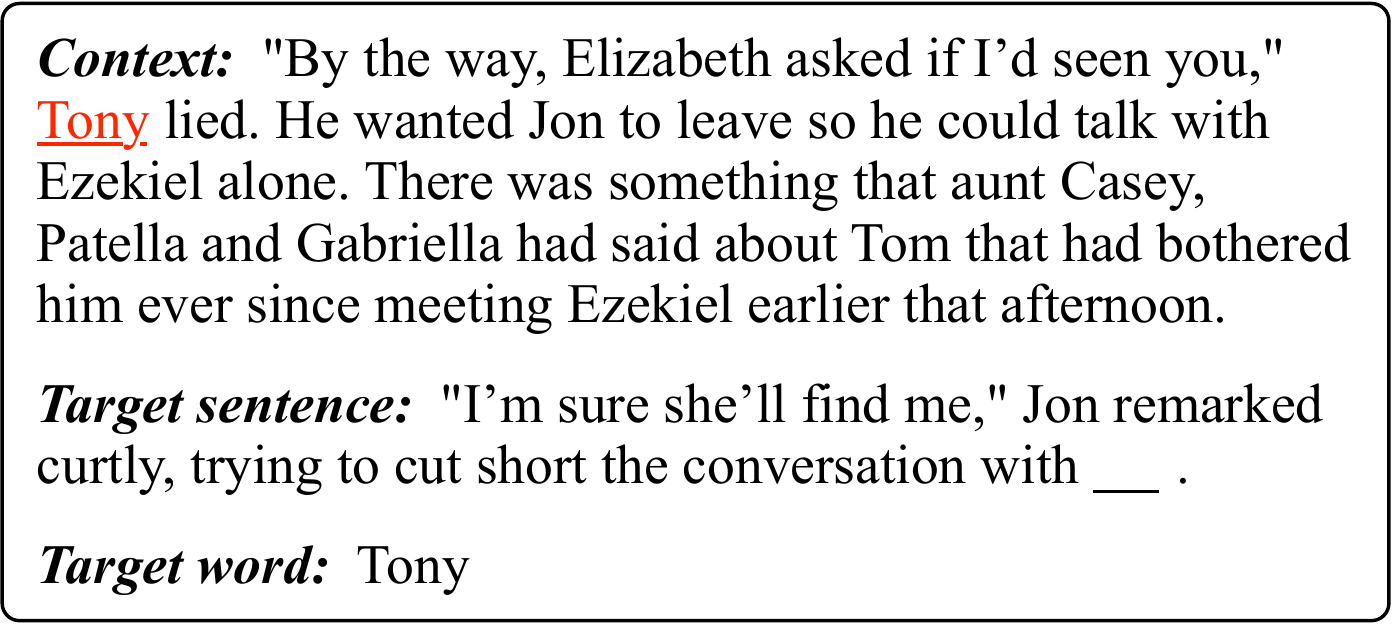}
	\caption{An example from the LAMBADA dataset.}
	\label{fig::lambada-example}
\end{figure}

\paragraph{The LAMBADA Task}
\citet{Paperno2016ACL} introduced the LAMBADA dataset, a specially designed language modeling task where each data point is a passage composed of a context (on average 4 to 5 sentences) and a target sentence, and the task is to guess the last word of the target sentence. The data comes from the BooksCorpus \cite{Zhu2015ICCV}, and is filtered by human subjects such that it is easy for humans to guess the target word when provided with the whole passage, but impossible to guess given only the target sentence. An example is shown in Figure \ref{fig::lambada-example}.

\citet{Paperno2016ACL} also reported the results of some standard language models on the task, which are extremely low as none of them reached an accuracy of 1\%, while a baseline of selecting a random capitalized word from passage gave an accuracy of 7.3\%, indicating the difficulty of the task.

Since then, \citet{Chu2017EACL} proposed to view LAMBADA as reading comprehension, with the context sentences as the context and the target sentence without the last word as the query. The model is then asked to select a word from the context as the answer. Despite the fact that models under this setup will decidedly fail on 19\% of the test cases where the target word is not in the context, doing so still greatly improved performance to 49\%. \citet{Dhingra2018NAACL} improved the number to 55.69\% by combining the Gated-Attention Reader \cite{Dhingra2017ACL} with a ``Coref-GRU'' layers (where both the previous token and the co-referent antecedent serve as the input to the current GRU cell). \citet{Hoang2018EMNLP} combined the Attention-Sum Reader \cite{Kadlec2016ACL} with a multi-task objective to track entities in the context, further improving performance to 59.23\%. Both these experiments proved the effectiveness of coreference knowledge in the task.

\begin{figure*}[!t]
	\centering
	\begin{subfigure}[b]{0.32\textwidth}
		\includegraphics[width=\linewidth]{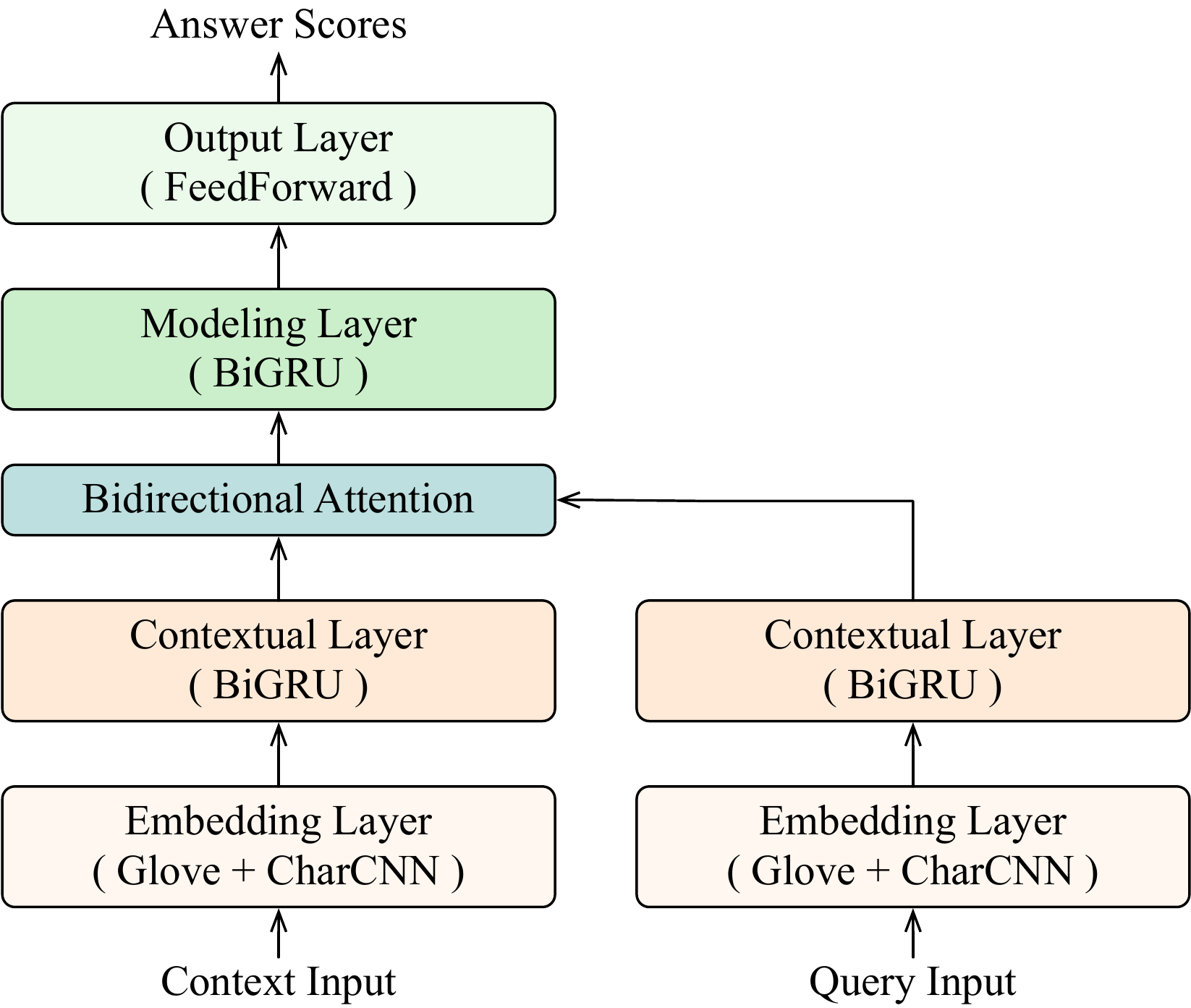}
		\caption{The baseline \textsc{BiDAF} model.}
		\label{fig::bidaf::base}
	\end{subfigure}
	\hfill
	\begin{subfigure}[b]{0.32\textwidth}
		\includegraphics[width=\linewidth]{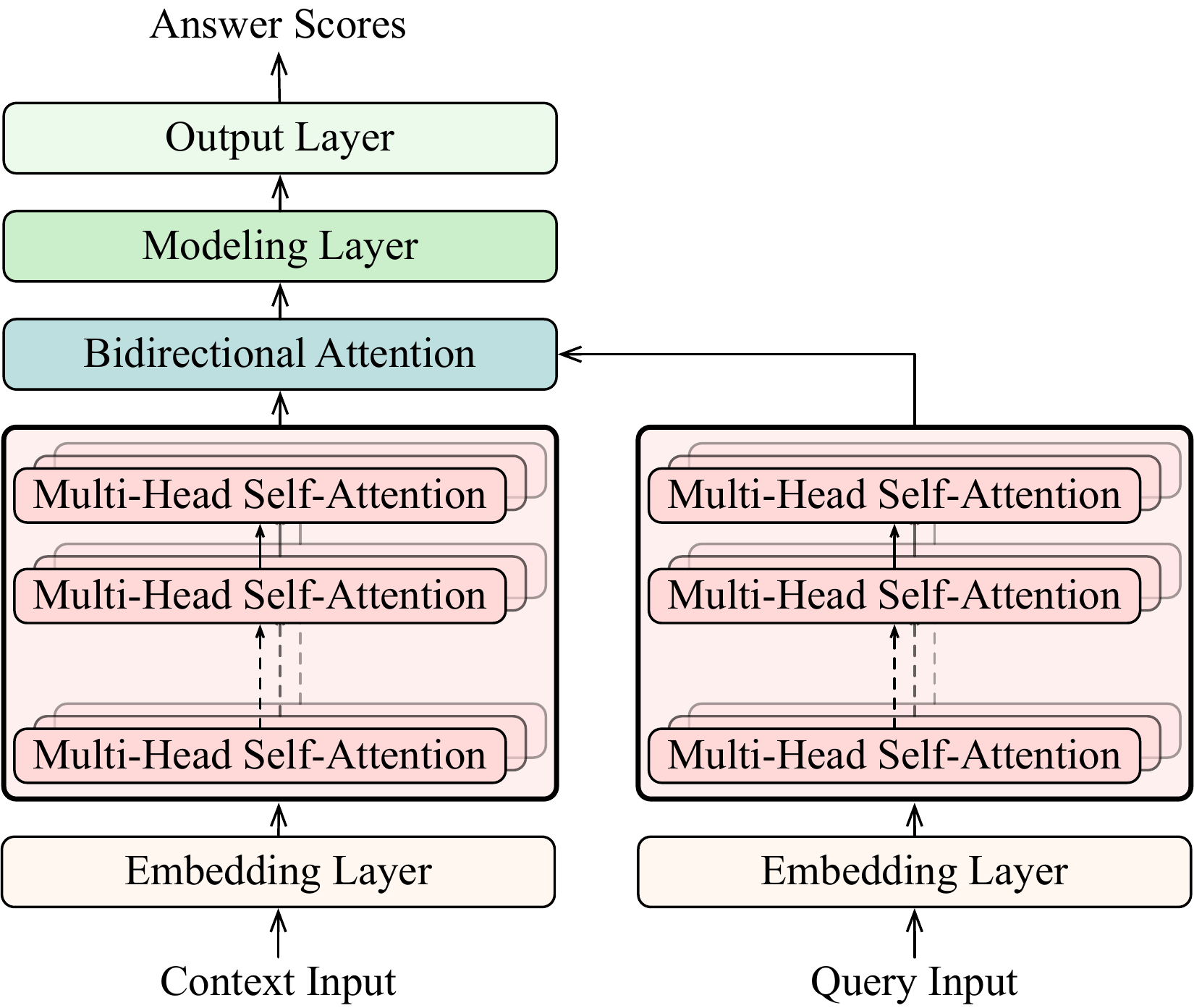}
		\caption{The \textsc{BiDAF-SA-Early} variant.}
		\label{fig::bidaf::early}
	\end{subfigure}
	\hfill
	\begin{subfigure}[b]{0.32\textwidth}
		\includegraphics[width=\linewidth]{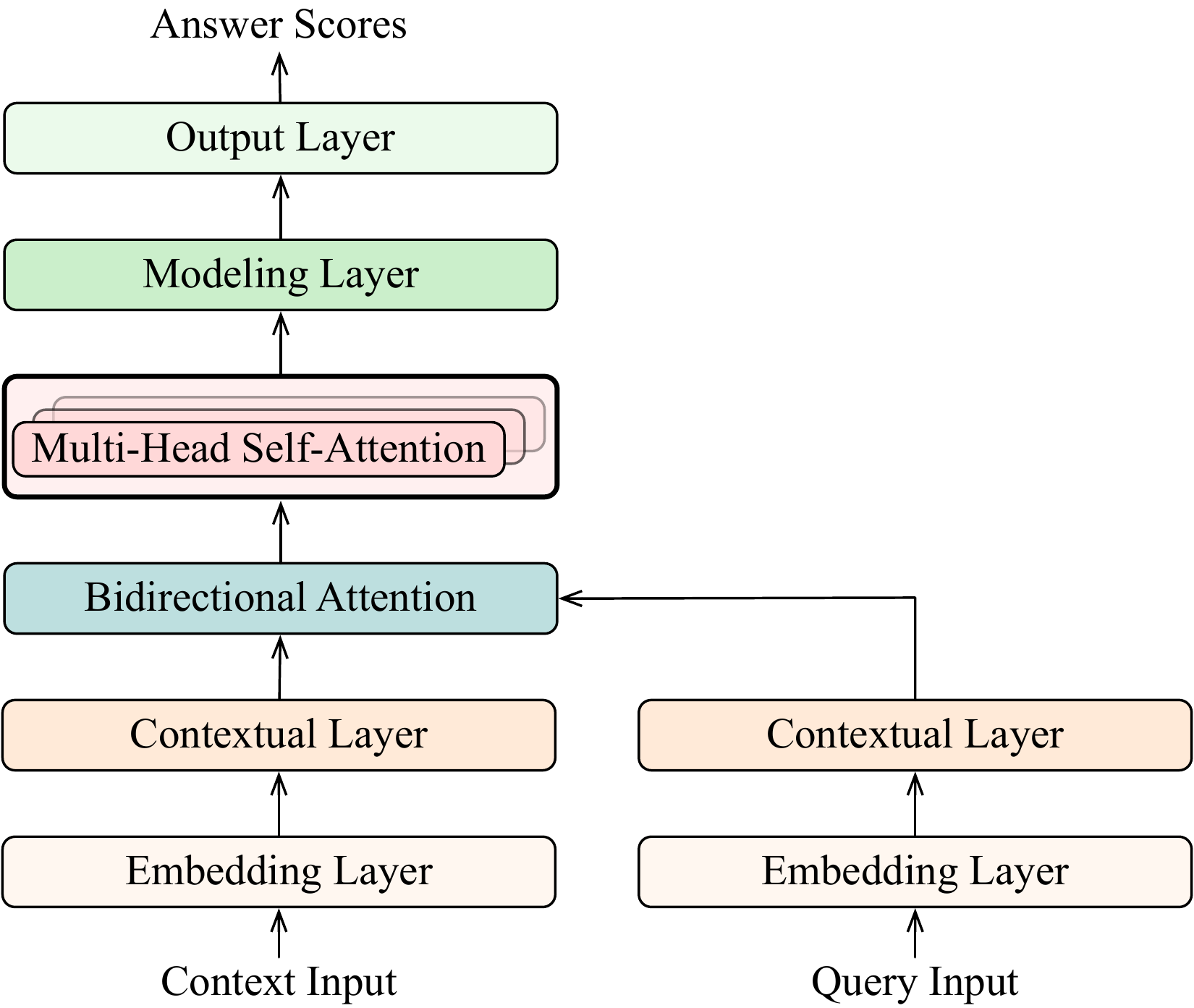}
		\caption{The \textsc{BiDAF-SA-Late} variant.}
		\label{fig::bidaf::late}
	\end{subfigure}
	\caption{The original \textsc{BiDAF} model \cite{Seo2017ICLR} that we use as a baseline in our experiment, and two variants with a self-attention encoder (in red) being added either as the contextual layer or after the bidirectional attention layer.}
	\label{fig::bidaf}
\end{figure*}

There have also been some efforts in applying Transformer-style models to the task. \citet{Dehghani2019ICLR} achieved 56.25\% with Universal Transformers. Most recently, \citet{Radford2019GPT2} reported 63.24\% with the largest GPT-2 model (1.5 billion parameters), setting the current state-of-the-art. Nonetheless, it is still far from the human performance of 86\% estimated by \citet{Chu2017EACL}.

\section{Method}
\label{sec::method}

\subsection{Task}
We adopt the same setting as most of the previous work on LAMBADA, that is, to view the task as reading comprehension. Formally, we concatenate all tokens in the context sentences to get the context input $\mathbf{x}=\{x_1\dots x_n\}$. We represent all but the last word from the target sentence as the query input $\mathbf{q}=\{q_1\dots q_m\}$, and the last word of the target sentence as the answer $a$.

The model computes a probability of being the correct answer for each word in the context $P(x_i | \mathbf{x}, \mathbf{q})$. Because the answer $a$ might occur multiple times in the context, at training time, we sum the probabilities of all correct tokens, and compute the loss as the negative log-likelihood of the summed probability:
\begin{equation}
\mathcal{L}_0 = -\log P(a | \mathbf{x}, \mathbf{q}) = -\log\sum_{i: x_i = a} P(x_i | \mathbf{x}, \mathbf{q})
\end{equation}
At test time, a pointer sum mechanism \cite{Kadlec2016ACL} is used to predict the word type with the highest summed probability among all distinct word types in the context.

\subsection{Model}

The aim of this paper is to test whether linguistic knowledge of semantic structures can be injected into an existing model via supervised self-attention, and whether the performance of such a model on the LAMBADA task can be matched with the large-scale pre-trained language models (i.e., GPT-2).\footnote{We take this route, rather than adding supervision to an existing GPT-2 model, because of the high computational cost of training such a large Transformer model from scratch.}

As discussed in Section \ref{sec::related}, a range of different reading comprehension models (i.e., Gated-Attention Reader, Attention-Sum Reader) have been tested in the previous work, and they all showed reasonably strong performance on the task \cite{Dhingra2018NAACL,Hoang2018EMNLP}. Therefore, we decide to start with a conventional reading comprehension model, and fuse into it a simpler and shallower stacked self-attention architecture (with fewer layers, fewer attention heads, and smaller hidden size compared to GPT-2). We choose another widely-used reading comprehension model, the BiDAF model \cite{Seo2017ICLR}, as our starting point, because BiDAF has consistently outperformed the aforementioned models in many reading comprehension benchmarks.

\paragraph{\textsc{BiDAF} Baseline}
The original \textsc{BiDAF} model, as illustrated in Figure \ref{fig::bidaf::base}, mainly consists of the following components:

\begin{enumerate}
	\item \textbf{Embedding Layer:} represents each token in the context and the query by a concatenation of Glove embeddings and Character-CNN embeddings.
	\item \textbf{Contextual Layer:} encodes the context sequence and the query sequence with a bidirectional-LSTM encoder.
	\item \textbf{Bidirectional Attention Layer:} computes both context-to-query and query-to-context attentions, which are then used to merge the query representations and the context representations to get query-aware vector representations for each context word.
	\item \textbf{Modeling Layer:} encodes the query-aware context representation with another bidirectional-LSTM encoder to capture the interaction among context words conditioned on the query.
	\item \textbf{Output Layer:} predicts the probability for each context word being the correct answer with a feed-forward layer followed by a softmax layer.
\end{enumerate}

\begin{figure*}[!t]
	\centering
	\begin{subfigure}[b]{0.3\textwidth}
		\includegraphics[width=\linewidth]{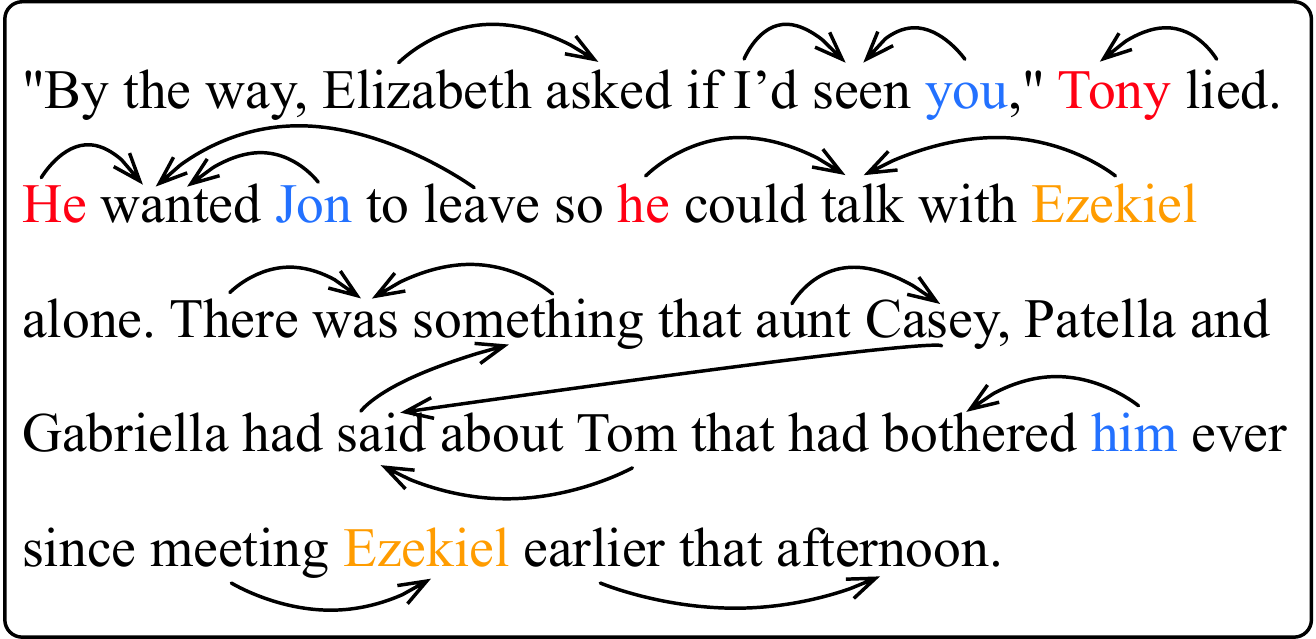}
		\caption{The dependency parses (arrows) and coreference chains (color-coded) of a context input (same as in Figure \ref{fig::lambada-example}), which are used to construct different auxiliary supervision signals, as shown on the right.}
		\label{fig::supv::preprocessing}
	\end{subfigure}
	\hfill
	\begin{subfigure}[b]{0.68\textwidth}
		\includegraphics[width=\linewidth]{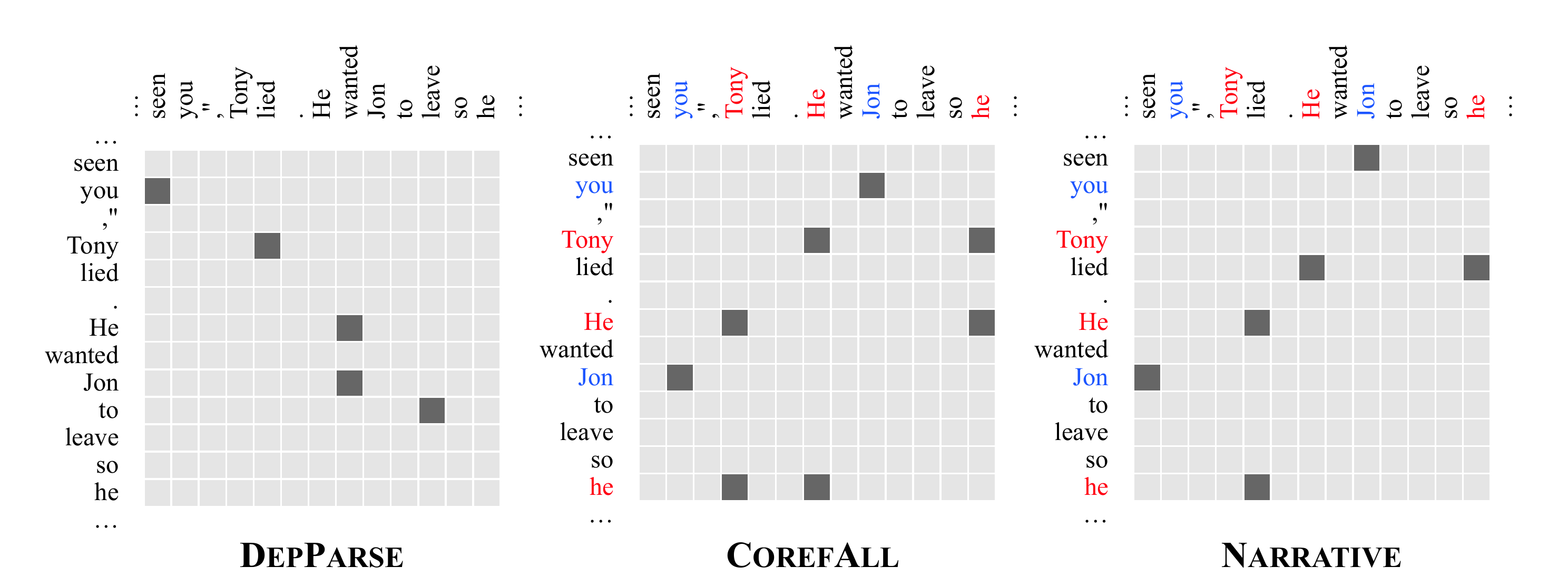}
		\caption{Examples of 3 different supervision types for self-attention (only showing part of the full matrices due to space limit). Light gray represents 0, and dark gray stands for 1.}
		\label{fig::supv::example}
	\end{subfigure}
	\caption{An example showing how we construct different types of supervision signals from pre-processed text input.}
	\label{fig::supv}
\end{figure*}

Our baseline model is mostly the same as the original \textsc{BiDAF} model, except for a few small changes: we substitute the LSTMs for GRUs; we add a Layer Normalization \cite{Ba2016LN} after the bidirectional attention layer and after the modeling layer to improve stability (see Section \ref{sec::exp}).

\paragraph{BiDAF with Self-Attention}

In order to inject semantic knowledge into the model via supervised attention, we need to fuse a stacked multi-head self-attention encoder into the \textsc{BiDAF} model. Intuitively, there are two options on where the self-attention encoder fits in:

\begin{enumerate}[label=\alph*)]
	\item Use the encoder to replace the \textbf{Contextual Layer}, as shown in Figure \ref{fig::bidaf::early}.
	
	This is inspired by the trend of using self-attention encoders to replace traditional RNN-based encoders in many NLP problems. Also, a common practice in using BERT  is to first encode raw input with BERT and then pass the output to higher-level task-specific layers, which is similar to what we do here. We name this variant \textsc{BiDAF-SA-Early}.
	
	\item Add the encoder after the \textbf{Bidirectional Attention Layer}, as shown in Figure \ref{fig::bidaf::late}.
	
	This is inspired by the \textsc{BiDAF++} model \cite{Clark2018ACL}, where a standard self-attention layer is added after the bidirectional attention layer to help reason over multiple paragraphs. Here we instead use multi-head self-attention, since applying auxiliary supervision on an attention layer with just one attention head leads to inferior performance in our preliminary experiments. We name this variant \textsc{BiDAF-SA-Late}.
	
\end{enumerate}

We also explore another variant that combines the two options, called the \textsc{BiDAF-SA-Both} model.

\subsection{Auxiliary Supervision for Self Attention}

Similar to \citet{Strubell2018EMNLP}, we want to apply auxiliary supervision on a self-attention encoder to guide the model to learn some specific linguistic structure. Our model receives context input, namely the passage, as well as query input, which is the target sentence minus the last word. We focus on investigating auxiliary supervision on the context input,  because the context input, with 4 to 5 sentences on average, should exhibit much richer linguistic structures than the query input, which is a single sentence.
To examine what kind of linguistic structures are beneficial to the problem, we experiment with 3 types of supervision signals:

\paragraph{Syntax Supervision:}
Given the dependency parses for each sentence in the context, we construct the target self-attention weights by putting a weight of 1 from each token to its syntactic head token, and 0 otherwise, as shown in the left column of Figure \ref{fig::supv::example}. We name this type of supervision \textbf{\textsc{DepParse}}.

This is similar to the auxiliary supervision used in \citet{Strubell2018EMNLP}, except that we have multiple sentences rather than just one sentence. If an attention head is trained with this syntax supervision, we constrain the self-attention window by sentence boundaries, that is, each token can only attend to other tokens in the same sentence, to make it easier for the model to approach the target self-attention weights.

\paragraph{Coreference Supervision:}
Given a list of coreference chains from the context (each coreference chain contains a set of mentions that refer to the same entity), we construct the target self-attention weights by putting a weight of 1 between each pair of mention heads in the same coreference chain, and 0 otherwise, as shown in the middle column of Figure \ref{fig::supv::example}. We name it \textbf{\textsc{CorefAll}}.

We also test other variants of coreference supervision, namely, guiding the head of each mention to only attend to the head of the most recent previous mention, or to the head of the immediately following mention. We refer to these two variants as \textbf{\textsc{CorefPrev}} and \textbf{\textsc{CorefNext}} respectively.

\paragraph{Narrative Supervision:}

Since the LAMBADA dataset is built from a corpus of novels, we hypothesize that narrative structures, that is, the sequence of events and their participants, could also be important for predicting the missing word. The interaction between the predicate and arguments of a single event is largely captured by the syntax supervision described above. Therefore, we combine the dependency parses and coreference chains to construct another type of self-attention targets that reflect higher-level narrative knowledge, as shown in the right column of Figure \ref{fig::supv::example}: For each event argument $a$, we put a weight of 1 between $a$ and all predicates that have an argument that co-refers with $a$. We name this supervision \textbf{\textsc{Narrative}}.

Note that while we require some extra information (i.e., dependency parses and coreference chains, as shown in Figure \ref{fig::supv::preprocessing}) to construct the auxiliary supervision signals, we do NOT rely on any gold annotations on either the training set or the test set. All the information can be obtained automatically from running existing NLP tools.
We discuss the pre-processing steps more in Section \ref{sec::exp}.

If any auxiliary supervision is applied to a self-attention head, we compute a loss from the supervision matrix $S\in \mathcal{R}^{n\times n}$ and the attention weights $A\in \mathcal{R}^{n\times n}$ as:
\begin{equation}
\mathcal{L}_s = \dfrac{1}{k} \sum_{i=1}^n \left[-\log\left(\sum_{j=1}^n A_{ij} * S_{ij}\right) * \sum_{j=1}^n S_{ij}\right]
\end{equation}
where $k$ is the number of rows in $S$ such that there is at least one non-zero element in the row.
To explain, for each token with at least one supervision target, we compute the negative log-likelihood loss against all of its supervision targets, and then get the mean value.

The model is trained in an end-to-end fashion, no matter whether any auxiliary supervision is applied. For a model combined with auxiliary supervisions $s_1,s_2,\dots$, the total loss being optimized is:
\begin{equation}
\mathcal{L} = \mathcal{L}_0 + \lambda * \sum_i \mathcal{L}_{s_i}
\end{equation}
where $\lambda$ is a hyper-parameter to balance between the final prediction loss and the auxiliary losses.

\section{Experimental Results}
\label{sec::exp}
\paragraph{Dataset \& Pre-processing}

When introducing the LAMBADA dataset, \citet{Paperno2016ACL} divided the BooksCorpus randomly into 2 partitions, and only applied the human subjects filtering process to the second half to create the development / test set, while leaving the first half raw data untouched to be the training set. With the reading comprehension setup introduced by \citet{Chu2017EACL}, they also constructed a new training set of \texttildelow 1.6M instances out of the original raw data, by requiring that the last word in the target sentence must exist in the context. Follow-up work \cite{Dhingra2018NAACL,Hoang2018EMNLP} further filtered the new training set by removing all instances where the target word is a stopword, leaving \texttildelow 700k instances. We follow the same setup here. A summary is shown in Table \ref{tab::data-summary}.

\begin{table}[!t]
	\small
	\centering
	\begin{tabular}{c c c c}
		\hline
		& \textsc{Train} & \textsc{Dev} & \textsc{Test} \\
		\hline
		Size & 709,568 & 4,869 & 5,153 \\
		\% Answer-in-context & 100\% & 82.4\% & 81.7\% \\
		Filtered by human subjects & No & Yes & Yes \\
		\hline
	\end{tabular}
	\caption{A brief summary of the LAMBADA dataset.}
	\label{tab::data-summary}
\end{table}

As discussed in Section \ref{sec::method}, we also need to get the dependency trees and coreference chains from the data in order to construct the target attention weights for auxiliary supervisions. We use the neural dependency parser and the statistical coreference system from Stanford CoreNLP toolkit \cite{Manning2014ACL} to pre-process the whole dataset. Further discussion on the choice of pre-processing alternatives will be in Section \ref{sec::analysis}.

\paragraph{Implementation Details}

We build our models and run all the experiments with AllenNLP \cite{Gardner2017AllenNLP}. For the baseline \textsc{BiDAF} model, we mostly follow the hyper-parameters of the original model: due to space limits, we provide a detailed description of hyper-parameter choices in the supplemental material.
We train the model for 10 epochs, and perform early-stopping when the validation accuracy does not increase for two consecutive epochs. We use validation accuracy to select the best epoch, from which then weights are then used for test set evaluation. 

For the multi-head self-attention encoders in the \textsc{BiDAF-SA-*} variants, we always use 4 attention heads per layer. For \textsc{BiDAF-SA-Early}, we include 4 layers in the stacked self-attention encoder, as preliminary studies show that using only 1 or 2 layers damages performance heavily, while using more than 4 layers gives no significant improvement. For \textsc{BiDAF-SA-Late}, we only add 1 multi-head self-attention layer, because again, preliminary results show no further gain of using 2 or more layers.
In some experiments, we also try replacing the embedding layer with the pre-trained ELMo embeddings \cite{Peters2018NAACL}.

We find that performance is very sensitive to the initial random state, possibly due to the fact that there is a large statistical discrepancy between the training set and the development / test sets (because the training set is not filtered by human subjects). We observed a similar effect when we re-ran existing models from the literature \cite{Dhingra2018NAACL,Hoang2018EMNLP}. Therefore, for each model variant, we train 4 different runs with different random seeds, and report the average and maximum performance (in parentheses in the following tables) across the 4 runs.

\paragraph{Results on LAMBADA}
We compare our method with the two best previous approaches that did not use large-scale pre-training language models \cite{Dhingra2018NAACL,Hoang2018EMNLP}, and GPT-2 \cite{Radford2018GPT}. Note that we do not compare to other major pre-trained LMs, because BERT \cite{Devlin2019NAACL} and its follow-ups like XLNet \cite{Yang2019xlnet} and RoBERTa \cite{Liu2019roberta} all used the BooksCorpus as part of the pre-training data. As the LAMBADA task is constructed from the BooksCorpus, BERT and other models would gain an unfair advantage on the task because all the test instances have been accessed by these models during pre-training.

We present our main results in Table \ref{tab::results-lambada}. Here we first focus on the \textsc{BiDAF-SA-Early} model and the \textsc{CorefAll} supervision, because intuitively, knowledge about coreference chains in the passage is likely to be the most beneficial factor for solving the task. Results from other variants are discussed in Section \ref{sec::analysis}.

\begin{table}[!t]
	\small
	\centering
	\begin{tabular}{l c}
		\hline
		Models & Accuracy (\%)  \\
		\hline
		\citet{Dhingra2018NAACL} & 55.69 \\
		\citet{Hoang2018EMNLP} & 59.23 \\
		\hline
		GPT-2 \cite{Radford2019GPT2} & 63.24 \\
		\hline
		\textsc{BiDAF} & 59.12 (59.54) \\
		\textsc{BiDAF-SA-Early} & 60.54 (60.88) \\
		\textsc{BiDAF-SA-Early} + \textsc{CorefAll} & 61.51 (61.94) \\
		\textsc{BiDAF-SA-Early} + ELMo & 61.38 (61.87) \\
		\textsc{BiDAF-SA-Early} + ELMo + \textsc{CorefAll} & \textbf{63.71 (64.62)} \\
		\hline
	\end{tabular}
	\caption{Accuracy on the LAMBADA \textsc{test} set: average accuracy across 4 runs, with max accuracy in parentheses.}
	\label{tab::results-lambada}
\end{table}

We see that our \textsc{BiDAF} baseline already performs similarly to the previous best results before GPT-2. Adding the \textsc{CorefAll} auxiliary supervision consistently improves accuracy, with or without ELMo embeddings, but we see larger improvement from \textsc{CorefAll} with ELMo embeddings (\texttildelow 2.3 points) compared to that without ELMo embeddings (\texttildelow 1 point). This confirms our hypothesis that the injection of semantic knowledge via supervised self-attention can be helpful in addition to recent pre-trained language models. Also, the fact that ELMo itself only brings less than 1 point of improvement without \textsc{CorefAll} emphasizes the difficulty of the task. With both ELMo embeddings and \textsc{CorefAll} supervision, we achieve an average accuracy of 63.71\% (and the best run achieves 64.62\%), outperforming the largest GPT-2 model. This is quite surprising, considering that our model only contains 2.6 million tunable parameters\footnote{Even if we take the frozen ELMo parameters into account, the total number of parameters is only 96 million.}, significantly smaller than the number of parameters in GPT-2 (1.5 billion).

\section{Analysis}
\label{sec::analysis}
In this section, we aim to understand why the coreference supervision helps, what is the best possible way to apply auxiliary supervision, and how different types of supervision signals compare.

\paragraph{Does pre-processing quality affect performance?}
The statistical coreference system from Stanford CoreNLP, from which we construct the supervision signals, is not the current best coreference model in terms of benchmark metrics. We also experimented with a more recent end-to-end neural coreference model \cite{Lee2017ACL}\footnote{We use a re-implementation from AllenNLP.}, with much higher benchmark scores, as supervision signal for our \textsc{BiDAF-SA-Early} + \textsc{CorefAll} model.
Surprisingly, this yields inferior performance (61.13\% on average, compared to 61.51\% with the Stanford coreference results).

\begin{figure}[!t]
	\centering
	\includegraphics[width=0.9\linewidth]{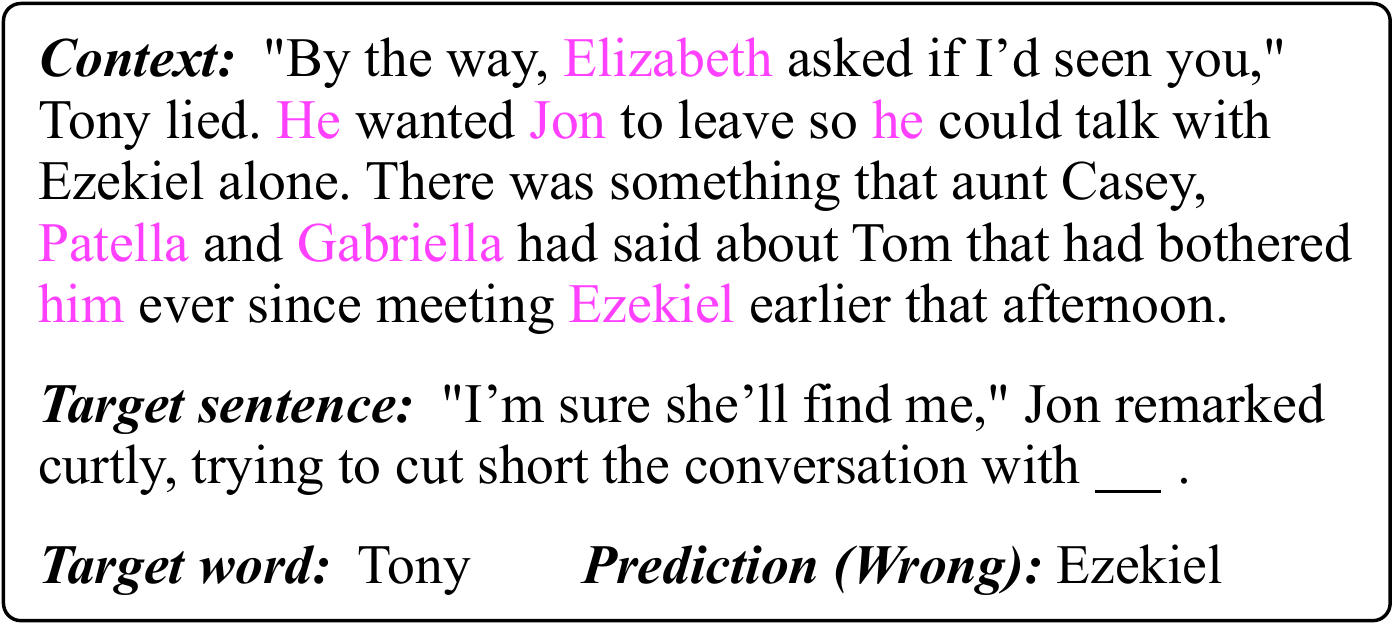}
	\caption{Does pre-processing quality affect performance? An example where a wrong coreference chain (color-coded) from a neural coreference system (which we do not use for our experiments) leads to the wrong prediction. A better coreference output from the Stanford coreference system on this example is shown in Figure \ref{fig::supv::preprocessing}.}
	\label{fig::wrong-prediction}
\end{figure}

We manually examined the output of both coreference systems on some data points, and found that the neural coreference system often produces highly erroneous output, possibly because it is highly optimized towards its news-centric training data, the OntoNotes dataset \cite{Hovy2006Ontonotes}, while LAMBADA consists of narrative texts. Figure \ref{fig::wrong-prediction} shows an example where a wrong coreference chain from the neural system leads to a wrong prediction. This is the same example as in Figure \ref{fig::supv::preprocessing}, which shows the coreference output from the Stanford system. In this example, it is hard to predict the right answer without knowing that ``you'' refers to ``Jon'' and ``he'' refers to ``Tony'', both of which are predicted correctly by the Stanford system.

This indicates that a better coreference signal could lead to even better results on the task. We leave it to future work, given some very recent work that further improved coreference performance \cite{Joshi2019spanbert}.

\paragraph{Does \textsc{CorefAll} really learn coreference knowledge?}
We want to know whether the improvement from \textsc{Coref\-All} supervision is because the supervision will actually allow the model to better learn coreference structures, or due to some unknown confounding factors. \citet{Chu2017EACL} manually analyzed 100 random instances from the LAMBADA \textsc{dev} set to identify the type of reasoning needed for humans to make the right prediction, and found that 21 out of the 100 instances require coreference resolution. We test our models on these 21 instances. To obtain a larger set of instances, we also compare the cases in the \textsc{dev} set where the target word is a noun to cases where the target is a pronoun, and we compare the cases where the target word is a \textsc{Person} to the cases where it is not a named entity (the most common named entity type is \textsc{Person}, with all other types occurring very rarely, so we focus on \textsc{Person}).

\begin{table}[!t]
	\small
	\centering
	\begin{tabular}{l c c c}
		\hline
		\textsc{dev} subset & \# & no supervision & with supervision \\
		\hline
		Require coref & 21 & 48.5 $\pm$ 4.1 & \textbf{62.0 $\pm$ 3.5} \\
		\hline
		Noun & 2,006 & 58.42 $\pm$ 0.32 & 59.48 $\pm$ 0.23 \\
		Pronoun & 2,138 & 72.31 $\pm$ 0.15 & \textbf{76.72 $\pm$ 0.53} \\
		\hline
		Not NE & 2,848 & 54.15 $\pm$ 0.21 & 54.90 $\pm$ 0.33 \\
		\textsc{Person} & 1,646 & 72.71 $\pm$ 0.08 & \textbf{77.95 $\pm$ 0.66} \\
		\hline
	\end{tabular}
	\caption{Does \textsc{CorefAll} learn coreference? Accuracy on some \textsc{dev} subsets, \textsc{BiDAF-SA-Early} + ELMo model with and without \textsc{CorefAll} supervision. We report average and  standard deviation across 4 runs.} 
	\label{tab::coref-analysis}
\end{table}

The results in Table \ref{tab::coref-analysis} show that not only does \textsc{CorefAll} supervision improve accuracy on the 21 instances manually classified as requiring coreference, it also strongly boosts performance on the ``Pronoun'' and ``PERSON'' subsets, in comparison to their ``Noun'' and ``Not NE'' counterparts. Though not a direct proof, this intuitively supports the claim that the auxiliary supervision does enable the model to better capture  coreference information, which is likely to help to reason particularly over pronouns and named entities.

\begin{table}[!t]
	\small
	\centering
	\begin{tabular}{l c}
		\hline
		Models & Accuracy (\%)  \\
		\hline
		\textsc{BiDAF-SA-Early} & 60.54 (60.88) \\
		\textsc{BiDAF-SA-Early} + \textsc{CorefAll} & 61.51 (61.94) \\
		\hline
		\textsc{BiDAF-SA-Late} & 59.48 (59.58) \\
		\textsc{BiDAF-SA-Late} + \textsc{CorefAll} & 61.19 (61.54) \\
		\hline
		\textsc{BiDAF-SA-Both} & 60.88 (61.27) \\
		\textsc{BiDAF-SA-Both} + \textsc{CorefAll} (early) & 61.72 (62.35) \\
		\textsc{BiDAF-SA-Both} + \textsc{CorefAll} (late) & 61.54 (61.67) \\
		\hline
	\end{tabular}
	\caption{Location of supervision: Accuracy (average of 4 runs, with max in parentheses) on the LAMBADA \textsc{test} set for different locations to fuse in the self-attention encoder and to apply auxiliary supervision.}
	\label{tab::early-vs-late-supv}
\end{table}

\paragraph{Where should the supervision be applied?}
Is it more beneficial to apply auxiliary supervision at an earlier stage of the model, i.e., at the contextual layer as in the \textsc{BiDAF-SA-Early} model, or at a later stage, i.e., after the bidirectional attention layer as in the \textsc{BiDAF-SA-Late} model? We compare performance using \textsc{CorefAll} supervision. Also, to disentangle the effect of architectural change, we experiment with the \textsc{BiDAF-SA-Both} model with supervision being added at different stages.

Table \ref{tab::early-vs-late-supv} shows that without supervision, \textsc{BiDAF-SA-early} offers much better results than \textsc{BiDAF-SA-late}. Although adding supervision to \textsc{BiDAF-SA-Late} leads to a larger relative improvement, applying supervision at an earlier stage still leads to a better absolute performance than doing so at a later stage, which is also confirmed by the numbers on \textsc{BiDAF-SA-Both}. This is not surprising, as intuitively coreference information about the context input should be beneficial to getting better query-aware context representations, rather than the other way around.\footnote{In all our experiments with \textsc{BiDAF-SA-Early}, we apply auxiliary supervision on the 3rd layer (out of all 4 layers). We also test applying supervision on other layers, and find the 3rd layer generally works the best, though the difference is not significant.}

\paragraph{Are other types of supervision also useful?}
We have so far focused on the \textsc{CorefAll} supervision. In Table \ref{tab::supv-type} we show the results of applying other types of auxiliary supervision. 

\begin{table}[!t]
	\small
	\centering
	\begin{tabular}{l c}
		\hline
		Models & Accuracy (\%)  \\
		\hline
		\textsc{BiDAF-SA-Early} & 60.54 (60.88) \\
		\textsc{BiDAF-SA-Early} + \textsc{CorefAll} & 61.51 (61.94) \\
		\hline
		\textsc{BiDAF-SA-Early} + \textsc{DepParse} &  61.06 (61.34) \\
		\textsc{BiDAF-SA-Early} + \textsc{CorefPrev} &  60.94 (61.71) \\
		\textsc{BiDAF-SA-Early} + \textsc{CorefNext} &  61.27 (61.63) \\
		\textsc{BiDAF-SA-Early} + \textsc{Narrative} & 61.86 (62.39) \\
		\hline
	\end{tabular}
	\caption{Supervision types: Accuracy on the LAMBADA \textsc{test} set (average of 4 runs, with max in parentheses) with different types of auxiliary supervision.}
	\label{tab::supv-type}
\end{table}

All other types of auxiliary supervision, except for \textsc{Narrative}, show inferior performance compared to \textsc{CorefAll}. This is as expected. As the LAMBADA task is specifically designed to require broader discourse context, intra-sentence syntactic structure (\textsc{DepParse}) should not play an important role. The \textsc{CorefPrev} and \textsc{CorefNext} variants of coreference information only provide guidance towards the immediately preceding or following mention in the same coreference chain. Such knowledge will fall short when the reasoning over a long coreference chain is crucial in making the prediction.

The \textsc{Narrative} supervision provides slightly better performance than \textsc{CorefAll}. This is also not surprising, as the \textsc{Narrative} signal is derived from both dependency parses and coreference chains. Theoretically, this type of supervision should capture useful linguistic structures from both \textsc{CorefAll}, which makes the main contribution to the performance improvement, and \textsc{DepParse}, which might offer some additional boost.\footnote{We also briefly test combining multiple supervision signals, and find that \textsc{CorefAll} + \textsc{Narrative} leads to slightly better performance than using \textsc{Narrative} alone, but the difference is not significant as well.}
We further verify the hypothesis by computing the agreement between the predictions of two models on the \textsc{dev} set, and find that on average, a run with \textsc{CorefAll} and a run with \textsc{Narrative} agree on 89.3\% of all \textsc{dev} instances, confirming that it is largely the coreference signal that leads to the performance improvement observed with the \textsc{Narrative} supervision.

\section{Conclusion}
\label{sec::conclusion}
In this paper we have investigated whether the injection of semantic knowledge into an existing model (\textsc{BiDAF}) via supervised self-attention can lead to better performance on tasks requiring complex and long-distance reasoning.
On the LAMBADA dataset, where the current best result from GPT-2 \cite{Radford2019GPT2} is still far below human performance, we show that a \textsc{BiDAF} model trained with coreference as auxiliary supervision achieves the new state-of-the-art, while requiring only a tiny fraction of the parameters in GPT-2. 
We further tested model variants to test where in a \textsc{BiDAF} model it is most useful to add a self-attention layer with supervision, and how different types of linguistic knowledge compared as supervision signal.

This paper takes a first step in explicitly using structural semantic knowledge to inform self-attention,  which leads to many interesting future directions. First, we want to test other types of linguistic knowledge, for example, semantic role labeling or AMR parsing \cite{Ogorman2018AMR}. We also want to see how our current approach can be applied to other tasks, for example, the new \textsc{Quoref} dataset \cite{quoref:19} that requires resolving coreference among entities to answer a question. In this paper, we extract supervision signals from existing NLP pipeline tools, and the signals are actually quite noisy (especially on coreference). It would be interesting to see whether such semantic structures can be learned jointly when pre-training language models, with some distant supervision (for example, in a Wikipedia document, tokens with links pointing to the same Wikipedia page should be considered as co-referential mentions).

\section*{Acknowledgments}

This research was supported by the DARPA AIDA program under AFRL grant FA8750-18-2-0017. We acknowledge the Texas Advanced Computing Center for providing grid resources that contributed to these results, and some results presented in this paper were obtained using the Chameleon testbed supported by the National Science Foundation. We would like to thank the anonymous reviewers for their valuable feedback.

\fontsize{9.0pt}{10.0pt} \selectfont
\bibliographystyle{aaai}
\bibliography{ref}

\appendix

\section{More on experimental details}
\vspace{10pt}
\begin{enumerate}
	\item
	Our baseline \textsc{BiDAF} model largely follows the hyper-parameters of the original model:

	\begin{itemize}
		\item We use the concatenation of 100d Glove embeddings and 100 1D filters each with a width of 5 for Character-CNN embeddings for the embedding layer (with a combined output dimension of 200).
		\item We use 1-layer BiGRU for the contextual layer, and 2-layer BiGRU for the modeling layer, each with a hidden size of 100.
		\item We apply a dropout with rate 0.1 on the character CNN, all BiGRU layers, and the feed-forward layer before the final prediction.
		\item We train with a batch size of 128, and the Adam optimizer with a learning rate of 0.001.
		\item During training, we maintain a moving average of all model weights with a exponential decay rate of 0.9999. At test time, the moving average is used instead of the raw weights.
	\end{itemize}

	\item
	For the self-attention layers in the \textsc{BiDAF-SA-*} variants:
	\begin{itemize}
		\item We use a total dimension of 200 for the projected keys, queries, and values respectively (which are then divided by the number of attention heads) in each multi-head self-attention layer, to maintain the same size on later layers as the baseline model.
		\item In each self-attention layer, we also apply a dropout with rate 0.1 on the attention sub-layer, the feed-forward sub-layer, and the residual connections.
	\end{itemize}

	\item
	In training the \textsc{BiDAF-SA-*} variants, we use almost the same setup as in training the baseline model, except that we also use the learning rate scheduler as described in original Transformer paper (Vaswani et al. 2017), with a warm-up step size of 8,000:
	\begin{equation*}
	lr = d_{model}^{-0.5}\cdot \min(step^{-0.5},\ step^{0.5}\cdot warmup^{-1.5})
	\end{equation*}

	\item
	For models trained with auxiliary supervision, we use a multiplier $\lambda=0.3$ for the auxiliary loss in Equation (3). We test different $\lambda$ values: 0.1, 0.3, 0.5, 0.7, 1.0, and find that 0.3 works the best.
	
	\item
	In experiments where we use the ELMo embeddings, we use a dropout with rate 0.2 and projected the embeddings down to 200d, in order to keep the same output dimension as the original embedding layer.

\end{enumerate}

\end{document}